\documentclass{article}

\usepackage{microtype}
\usepackage{graphicx}
\usepackage{subfigure}
\usepackage{booktabs} %

\usepackage{hyperref}

\usepackage[accepted]{icml2025}

\usepackage{amsmath}
\usepackage{amssymb}
\usepackage{mathtools}
\usepackage{amsthm}

\usepackage{multirow}
\usepackage{booktabs}
\usepackage{makecell}
\usepackage{hhline}

\usepackage[capitalize,noabbrev]{cleveref}

\theoremstyle{plain}

\theoremstyle{definition}

\theoremstyle{remark}

\usepackage[textsize=tiny]{todonotes}

\icmltitlerunning{Learning Hamiltonian Dynamics with Bayesian Data Assimilation}

\begin{document}

\twocolumn[

\icmltitle{Learning Hamiltonian Dynamics with Bayesian Data Assimilation}

\icmlsetsymbol{equal}{*}

\begin{icmlauthorlist}
\icmlauthor{Taehyeun Kim}{equal,umich}
\icmlauthor{Tae-Geun Kim}{equal,yonsei}
\icmlauthor{Anouck Girard}{umich}
\icmlauthor{Ilya Kolmanovsky}{umich}
\end{icmlauthorlist}

\icmlaffiliation{umich}{Department of Aerospace Engineering, University of Michigan, Ann Arbor, MI, USA}
\icmlaffiliation{yonsei}{Department of Physics, Yonsei University, Seoul, Republic of Korea}

\icmlcorrespondingauthor{Taehyeun Kim}{taehyeun@umich.edu}
\icmlcorrespondingauthor{Ilya Kolmanovsky}{ilya@umich.edu}

\icmlkeywords{Machine Learning, ICML}

\vskip 0.3in
]

\printAffiliationsAndNotice{\icmlEqualContribution} %

\begin{abstract}
In this paper, we develop a neural network-based approach for time-series prediction in unknown Hamiltonian dynamical systems. Our approach leverages a surrogate model and learns the system dynamics using generalized coordinates (positions) and their conjugate momenta while preserving a constant Hamiltonian. To further enhance long-term prediction accuracy, we introduce an Autoregressive Hamiltonian Neural Network, which incorporates autoregressive prediction errors into the training objective. Additionally, we employ Bayesian data assimilation to refine predictions in real-time using online measurement data. Numerical experiments on a spring-mass system and highly elliptic orbits under gravitational perturbations demonstrate the effectiveness of the proposed method, highlighting its potential for accurate and robust long-term predictions.

\end{abstract}

\section{Introduction} \label{sec:intro}
A classical approach to system modeling is based on the first principles, see, e.g., industrial robotic arms~\cite{zada2016mathematical}, underwater vehicles~\cite{leonard1998mechanics}, unmanned aerial vehicle~\cite{mohamed2022hamiltonian}, and spacecraft~\cite{gurfil2016celestial} for specific examples. 

\begin{figure}[!t]
\vskip 0.2in
\begin{center}
\centerline{\includegraphics[width=3 in]{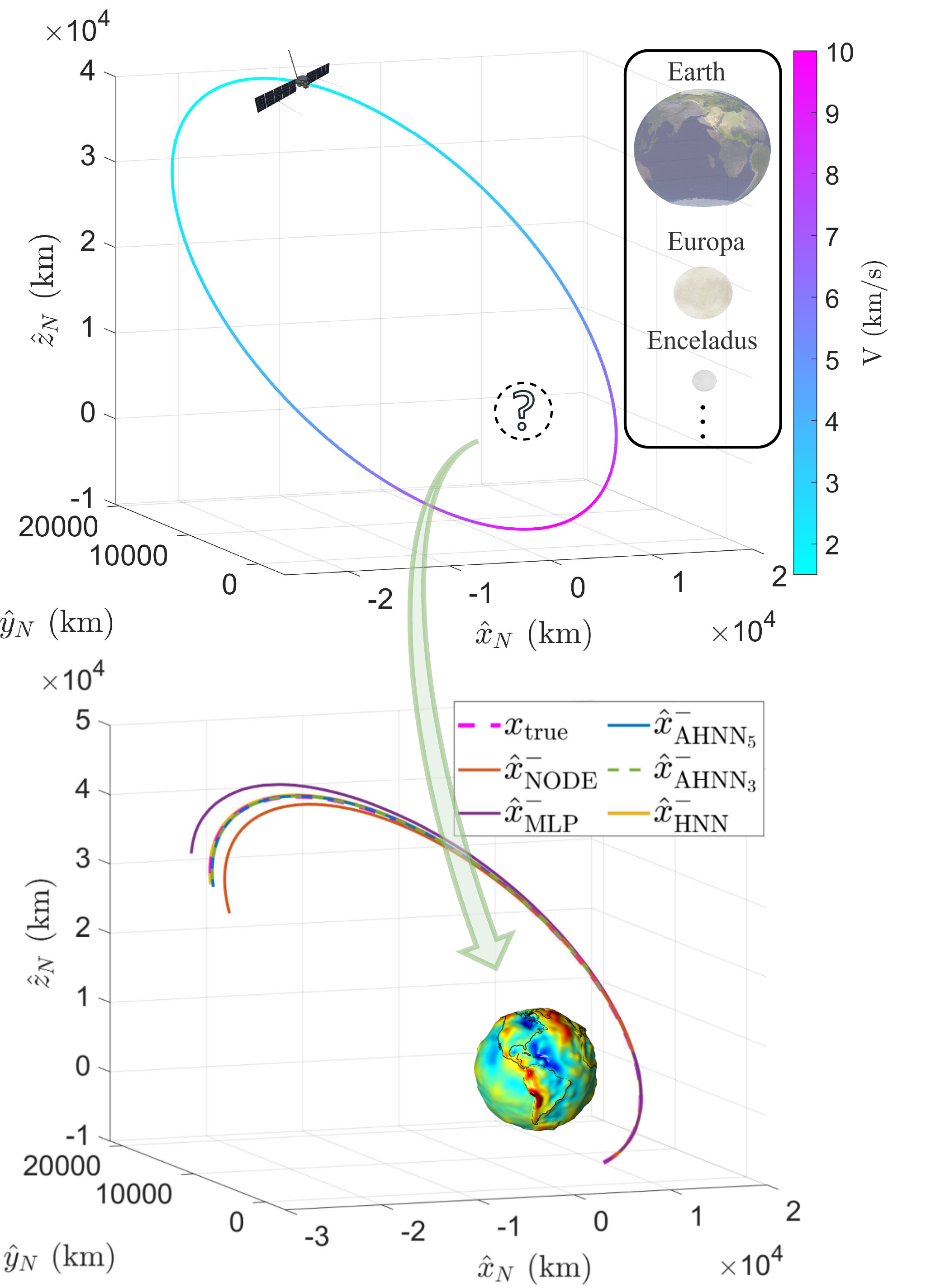}}
\caption{Learning unknown dynamics from spacecraft position and velocity data. The model performance is compared for a highly elliptical orbit, which poses significant challenges when the spacecraft is near the primary body due to rapid velocity changes, with speed variations up to an order of magnitude.}
\label{fig:elliptic_orbit}
\end{center}
\vskip -0.3in
\end{figure}

A classical problem in space flight mechanics which goes back to Kepler and Newton is that of orbit determination which involves position and velocity determination and prediction of various objects such as planets, asteroids, spacecraft, and space debris. 
The effectiveness of orbit determination depends on accuracy of dynamical models, estimators, and processing algorithms~\cite{selvan2023precise}. Among these, the choice of dynamical model can influence significantly the prediction accuracy in orbit determination problems~\cite{pastor2021initial}. 
Implementing missions for identifying parameters in such models requires considerable resources and specially designed experiments. In particular, the NASA Gravity Recovery and Climate Experiment (GRACE) mission~\cite{watkins2015improved} involved measurements of the relative motion of two spacecraft to accurately map Earth's gravitational field, illustrated as a colored contour map in Figure~\ref{fig:elliptic_orbit}.  

Recent advances in Machine Learning (ML) have enabled more accurate and efficient solutions to orbit determination problems, particularly in predicting object trajectories~\cite{caldas2024machine,kazemi2024orbit}. Early approaches employed conventional multilayer perceptron (MLP) models for circular orbit determination~\cite{peng2018artificial,salleh2022modeling}. However, these models exhibited overfitting issues~\cite{peng2019comparative} and struggled with long-term predictions, primarily due to their inability to account for the temporal relationships between consecutive states~\cite{greydanus2019hamiltonian}. While Long Short Term Memory (LSTM)-based predictors were later introduced to process sequential data for orbit determination~\cite{zhou2023lstm}, their real-time application was limited by the need to accumulate a predetermined window of sequential data before initial state prediction.

The evolution of ML approaches led to methods that better accommodate temporal continuity, such as neural ordinary differential equations (NODE)~\cite{chen2018neural} and Physics Informed Neural Networks (PINN)~\cite{raissi2019physics}. PINN implementations have shown promising results across various orbit families by leveraging astrodynamics domain knowledge~\cite{scorsoglio2023physic,loshelder2025orbit}. However, these approaches differ from our focus in this paper on developing methods that can be applied effectively without specific domain expertise. NODE improved OD performance by integrating state derivatives over time to connect sequential states~\cite{subramanian2023orbit}. Nevertheless, long-term predictions often deviated from actual trajectories because the trained models failed to capture exact conservation laws - a limitation that Hamiltonian Neural Networks (HNN) meliorate~\cite{greydanus2019hamiltonian}. While a HNN preserves the Hamiltonian structure of the model, its neural network-based predictions inevitably accumulate errors over time when applied autoregressively, causing predictions to drift away from true trajectories.

In particular, even with the true initial state, the accumulated prediction errors over time cause the predicted trajectory to diverge from the true trajectory. Furthermore, the true state at the current time instant may be uncertain and the predicted trajectory starting from an inaccurate initial state can easily diverge from the true state trajectory. Thus, when the observation telemetry data are available, it is beneficial to assimilate these data so that to improve prediction accuracy. We integrate HNN with the Unscented Kalman Filter (UKF)~\cite{julier1997new,wan2000unscented}, which is used for state estimation of a nonlinear system incorporating the observation data.

Our major contributions in addressing the above challenges are as follows:
\begin{itemize}
    \item (Algorithm) We propose a loss function for HNN training which enables to effectively learn unknown dynamics, including model structure and coefficients, based on position and velocity data.
    
    \item (Algorithm) We propose an approach to further improve HNN prediction accuracy by exploiting an autoregressive form of the loss function for training.
    
    \item (Algorithm) We integrate HNN with the Unscented Kalman Filter, which assimilates real-time measurements to improve prediction accuracy and enhances prediction stability; notably such an approach can also provide uncertainty highly desirable in orbit determination problems.
    
    \item (Validation) We demonstrate the effectiveness of HNN-based UKF in spacecraft orbit determination problems on highly elliptic orbits. For illustrative purposes, we also report the results from a simpler mass-spring example.
\end{itemize}

The rest of the paper is organized as follows. Section~\ref{sec:rel_work} surveys related work and highlights our contributions. Section~\ref{sec:method} describes the problem formulations and our proposed method. Section~\ref{sec:dyn} describes dynamic models of a simple mass-spring system
and of the Two-Body Problem (2BP) in space flight mechanics with additional gravitational perturbations. Section~\ref{sec:exp} presents the experimental results on a mass-spring system (1D motion) 
and spacecraft OD on highly elliptic orbits in 2BP with gravitational perturbations (3D motion). Section~\ref{sec:conclusion} presents conclusions and future research directions.

\section{Related Work} \label{sec:rel_work} 
Several Machine Learning approaches have been proposed for learning differential equations from data. These include Gaussian processes (GP) to learn the coefficients of partial differential equations~\cite{raissi2018hidden,raissi2018multistep} and state estimation application using linear system model~\cite{kuper2022numerical}. Following these developments, neural networks (NN) emerged as powerful tools for learning equations of motion from data.

Early NN-based models, including Residual networks and recurrent neural networks, focused on predicting discrete state sequences through Euler integration of continuous dynamics~\cite{lu2018beyond,ruthotto2020deep}.
Neural ordinary differential equations (NODE) advanced this approach by learning vector fields~\cite{chen2018neural} and employing numerical ODE solvers, such as the Runge-Kutta method, for continuous state transformation. 
Building on these developments, Hamiltonian Neural Networks (HNN)~\cite{greydanus2019hamiltonian} were proposed which maintained the structure of the Hamiltonian equations of for the model; they were also integrated~\cite{galioto2020bayesian}. However, these networks were primarily trained using the difference between predicted and true state derivatives.

Despite the emerging interest in HNN for various applications, HNN remains unexplored in the following scenario that incorporates 1) training NN models without accurate knowledge of true state derivative, 2) uncertainty quantification of state prediction, and 3) application to spacecraft orbit determination for highly elliptic Molniya orbits.

\section{Method} \label{sec:method}
Our first goal is to predict a sequence of future states of Hamiltonian systems over time, given their current state. With observations, the next goal is to estimate a sequence of future states and corresponding uncertainties, given the best initial guess. The Neural Network (NN)-based model $F_\theta(\cdot)$ learns an ``unknown'' discrete-time dynamics model using a data set $\mathcal{D}$, which consists of positions and momenta.

\subsection{Hamiltonian Neural Network}
The standard Hamiltonian Neural Network (HNN)~\cite{greydanus2019hamiltonian} predicts the future state of a mechanical system trajectories of which satisfy the Hamilton's equations~\citep{hamilton1833general},
\begin{equation}
\frac{{\rm d}q}{{\rm d}t} = \frac{\partial \mathcal{H}}{\partial p}, \quad \frac{{\rm d}p}{{\rm d}t} = -\frac{\partial \mathcal{H}}{\partial q},
\end{equation}
where $q \in \mathbb{R}^{n}$ and $p \in \mathbb{R}^{n}$ denote the generalized coordinates and conjugate momenta, respectively, and $\mathcal{H}$ is the Hamiltonian of the mechanical system. The neural network (NN) is used to represent a parameterized Hamiltonian function $\mathcal{H}_\theta: \mathbb{R}^{2n} \to \mathbb{R}$, where $\theta$ are NN parameters determined by minimizing the following $L_2$ loss function: $\mathcal{L}_\text{HNN} (q,p;\theta)=\left\|\frac{\partial \mathcal{H}_\theta}{\partial p}-\frac{{\rm d} q}{{\rm d} t}\right\|^2+\left\|\frac{\partial \mathcal{H}_\theta}{\partial q}+\frac{{\rm d} p}{{\rm d} t}\right\|^2 .$
This standard HNN approximately preserves the symplectic structure of the system and improves prediction stability by propagating states consistently with the expectations from Hamiltonian mechanics (energy conservation, volume preservation, recurrence, etc.).
While this approach has shown to be effective~\cite{greydanus2019hamiltonian} when state derivatives $(\dot{q},\dot{p})$ are available in the data-driven setting, accurately measuring these derivatives is often challenging in practical applications. In particular, $\dot{p}$ depends on the acceleration measurements, which can be noisy.

In order to address this issue, we propose the following modification. Suppose a training dataset $\mathcal{D} = \left\{ (x_k,x_{k+1}) \right\}_{k=1}^{N_{\mathcal{D}}}$, where $x_k = (q_k, p_k) \in \mathbb{R}^{2n}$, consisting of states along trajectories, is available. If $x_k$ and $x_{k+1}$ are two states in sequence along the same trajectory, we can use the following loss function to assess model accuracy over one step
\begin{equation} \label{eq:hnn1_loss}
\mathcal{L}_\text{HNN}^{+} (x_k , x_{k+1};\theta) = d(F_\theta (x_{k}), x_{k+1}),
\end{equation}
where $d:\mathbb{R}^{2n}\times \mathbb{R}^{2n} \to \mathbb{R}_{\geq 0}$ denotes an arbitrary metric function and
\begin{equation} \label{eq:hnn_pred}
F_\theta (x_{k}) = x_k + \int_{t_k}^{t_{k+1}} \mathbf{J}\nabla \mathcal{H}_\theta \, dt, \quad \mathbf{J} = \begin{bmatrix} \mathbf{0}_n & \mathbf{I}_n \\ -\mathbf{I}_n & \mathbf{0}_n \end{bmatrix}.
\end{equation}
Here, $\nabla \mathcal{H}_\theta = \left[\frac{\partial \mathcal{H}_\theta}{\partial q}, \frac{\partial \mathcal{H}_\theta}{\partial p} \right]^{\sf T}$ denotes the gradient of Hamiltonian, and $\mathbf{J}$ represents the canonical symplectic matrix, which consists of null $\mathbf{0}_n$ and identity $\mathbf{I}_n$ matrices.
Then, we compute and optimize the mean of the loss function values in~\eqref{eq:hnn1_loss} over data points.
The product $\mathbf{J}\nabla \mathcal{H}_\theta$ compactly represents Hamilton's equations in vector form, representing the flow of both generalized coordinates and momenta. Equation~\eqref{eq:hnn_pred} represents the integration of the ODEs, which is performed by a dedicated ODE solver. 

\begin{figure}[ht!]
\vskip 0.1in
\begin{center}
\centerline{\includegraphics[width=2.5 in]{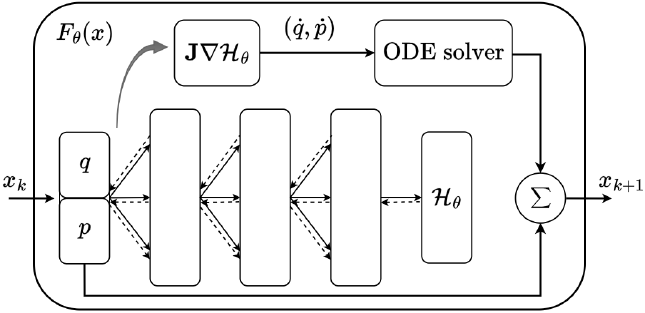}}
\caption{Illustration of the Hamiltonian Neural Network in~\eqref{eq:hnn_pred}.}
\label{fig:hnn}
\end{center}
\vskip -0.1in
\end{figure}

We use explicit ODE solvers for the integration to facilitate automatic differentiation for backpropagation. The schematics of the approach are shown in Figure~\ref{fig:hnn}.

\begin{figure*}[ht!]
\vskip 0.1in
\begin{center}
\centerline{\includegraphics[width=5 in]{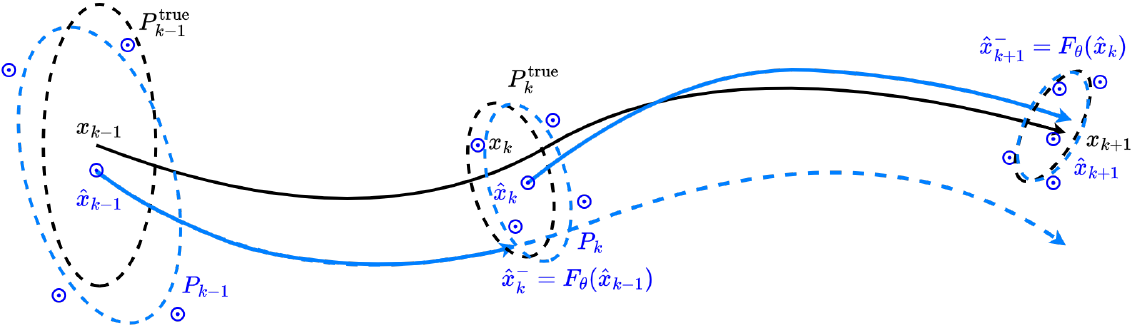}}
\caption{Illustration of the unscented transform for estimated mean and covariance propagation (blue) and true mean and covariance (black).}
\label{fig:hnn_ukf}
\end{center}
\vskip -0.1in
\end{figure*}
Our numerical experiments indicate that the HNN performs well on low-dimensional problems, but not on high-dimensional problems, e.g., $n\geq 3$. For spacecraft orbit determination and prediction, in particular, we have to deal with six dimensional state space. Furthermore, existing HNN approaches may lack accuracy for highly elliptic orbits with large variations in altitude and velocity. To improve long-term prediction accuracy, we introduce the Autoregressive HNN (AHNN) loss function which aims to enhance autoregressive prediction capabilities:
\begin{equation} \label{eq:loss_ahnn}
\mathcal{L}_{\text{AHNN}}(x_k, (x_{k+i})_{i=1}^{W};\theta) = \frac{1}{W}\sum_{i=1}^W d( \hat{x}_{k+i}, x_{k + i}  )
\end{equation}
where $\hat{x}_{k+i} = \overbrace{F_\theta \circ F_\theta \circ \cdots \circ F_\theta}^i (x_k)$ denotes the recursively predicted states based on the previous prediction and $(x_{k+i})_{i=1}^{W}$ is the sequence of true state values corresponding to the window size $W$, which is the number of autoregressive steps included in the loss computation. While the increase in $W$ enhances the robustness of AHNN with respect to accumulated prediction errors, it introduces computational overhead during backpropagation and complicates training. In this paper, we use Huber loss~\cite{huber1992robust} because of its robustness to extreme data in the training set and its everywhere differentiability properties.

\subsection{State Estimator}
\subsubsection{Conventional Kalman Filter}
The Kalman filter (KF)~\cite{kalman1960new} estimates the state of a discrete-time nonlinear system from observation data, i.e.,
\begin{align}
    x_{k+1}& = F_\theta(x_k)+\omega_k, \label{eq:dyn_model} \\
    y_{k}& = H(x_k)+\eta_k,  \label{eq:obs_model}
\end{align}
where $x_k\in \mathbb{R}^{2n},\; y_k \in \mathbb{R}^{n_o}$ denote the unobserved system state and the observation data, respectively, and $F_\theta:\mathbb{R}^{2n}\to \mathbb{R}^{2n}, H:\mathbb{R}^{2n}\to \mathbb{R}^{n_o}$ are the system dynamics model and observation model (such as distance/angle measurements by ground stations or distance to GPS satellites), respectively. In this work, we assume $H(\cdot)$ is known. The process noise $\omega_k \sim \mathcal{N}(0,\Sigma_\omega)$ comes from the uncertainty in the system model, and the observation noise $\eta_k \sim \mathcal{N}(0,\Sigma_\eta)$ comes from the uncertainty in the sensor model.

We assume the prior state estimate and measurement are Gaussian random variables (GRVs), i.e., $x_{k} \sim \mathcal{N}(\hat{x}_k, P_k)$ and $y_k \sim \mathcal{N}(H({x}_k), P^{yy}_{k})$, where $\hat{x}_k$, $P_k$, and $P^{yy}_{k}$ are the mean state vector, the state error covariance, and the measurement covariance at a time instant $t_k$, respectively.

The Kalman filter consists of two steps: the prediction and correction.
The prediction step, first, integrates an estimated state $\hat{x}_k$ and covariance $P_k$ from $t_k$ to $t_{k+1}$.
\begin{align} \label{eq:kf_pred}
    \hat{x}^-_{k+1} & = \mathbb{E} [F_\theta (x_k)|\mathcal{Y}_{k}], \\
    P^-_{k+1} & = {\rm Cov} [\hat{x}_{k+1}, \hat{x}_{k+1} |\mathcal{Y}_{k}], \nonumber 
\end{align} 
where $\mathcal{N}(x_k;\hat{x}_k,P_k)$ is the Gaussian probability distribution function of $x_k$ given $\hat{x}_k$ and $P_k$, $\delta x_{k+1}=F(x_{k})-\hat{x}^-_{k+1}$ is the state residual, and $\mathcal{Y}_{k}=\{y_0,y_1,y_2,\ldots, y_{k}\}$ is a set of measurement until the time instant $t_{k+1}$. Note that the superscript, $-$, indicates the predicted value before measurement data assimilation.
Using the measurements, the measurement update step then corrects the state estimate as 
\begin{align} \label{eq:kf_correct}    
    \hat{x}_{k+1} &= \hat{x}^-_{k+1} + {P}^{xy}_{k+1} ({P}^{yy}_{k+1})^{-1} \delta y_{k+1}, \\
    P_{k+1} &= P^-_{k+1} - {P}^{xy}_{k+1} ({P}^{yy}_{k+1})^{-1} ({P}^{xy}_{k+1})^{\sf T}, \nonumber
\end{align}
where $\delta y_{k+1}=y_{k+1} - \hat{y}^-_{k+1}$ is the measurement residual,
\begin{align} \label{eq:kf_meas_statistics}    
    \hat{y}^-_{k+1} &= \mathbb{E} [ H(x_{k+1})|\mathcal{Y}_{k}], \\
    {P}^{xy}_{k+1} &= {\rm Cov} [ {x}_{k+1}, {y}_{k+1}|\mathcal{Y}_{k}], \nonumber \\
     {P}^{yy}_{k+1} &= {\rm Cov} [ {y}_{k+1}, {y}_{k+1} |\mathcal{Y}_{k}]. \nonumber 
\end{align} 
    
\subsubsection{Unscented Kalman Filter}
The Unscented Kalman Filter (UKF) is grounded in the unscented transformation (UT), which is an approach to quantify the statistics of an output of a nonlinear function of a random variable~\cite{julier1997new,wan2000unscented}. Considering the current state estimate, $\hat{x}_k$, we generate a sigma point set $\mathcal{X}$, which consists of $2L+1$ sigma points associated with the random variable dimension $L=2n$, i.e.,
\begin{align} \label{eq:def_sig_pts}
    \mathcal{X}^{(0)}_k & = \hat{x}_k, \\
    \mathcal{X}^{(i)}_k & = \hat{x}_k+\left(\sqrt{(L+\lambda)P_{k}}\right)^{(i)}, \nonumber\\
    \mathcal{X}^{(i+L)}_k & = \hat{x}_k - \left( \sqrt{(L+\lambda)P_{k}} \right)^{(i)}, i=1,\ldots,L, \nonumber\\
    W^{(0)}_{m} & = \lambda/ (L + \lambda), \nonumber\\
    W^{(0)}_{P} & = \lambda/ (L + \lambda) + (1-\alpha^2 + \beta), \nonumber\\
    W^{(i)}_{m} &= W^{(i)}_{P} = \lambda/ 2(L + \lambda), \nonumber
\end{align}
where $\mathcal{X}^{(i)}_k$ is $i$-th sigma points in $\mathcal{X}$ at time instant $t_k$, $\lambda = \alpha (L+\kappa)-L$, and $\left(\sqrt{(L+\lambda)P }\right)^{(i)}$ denotes the $i$-th row of the matrix square root. The sigma points generation and weighting process is controlled by adjusting tuning parameters ($\alpha,\beta,\kappa$). Here, $\alpha$ determines how widely the sigma points $\mathcal{X}^{(i)}_k$ are spread around the state estimate $\hat{x}_k$, while $\beta$ incorporates prior knowledge of the state distribution, with $\beta=2$ being optimal for Gaussian distributions~\cite{wan2000unscented}.

\begin{algorithm} [!t] %
   \caption{NN-UKF}
   \label{alg:hnn_ukf}
\begin{algorithmic}
   \STATE {\bfseries Input:} state estimate $\hat{x}_k$ and covariance $P_k$ for $k\in\mathbb{Z}_{\geq 0}$
   \STATE Create sigma points: 
   $$\mathcal{X}_k = \begin{bmatrix}
       \hat{x}_k\\ 
       \hat{x}_k + \left(\sqrt{(L+\lambda)P_{k}}\right)^{(i)}\\ 
       \hat{x}_k - \left(\sqrt{(L+\lambda)P_{k}}\right)^{(i)} 
   \end{bmatrix},i=1,\cdots,L$$
   \STATE Predict sigma points using NN-based transformation:
   \begin{align}
       \mathcal{X}^-_{k+1} & = F_{\theta}(\mathcal{X}_{k}) \nonumber \\
       \hat{x}^-_{k+1} & = \sum_{i=0}^{2L}W^{(i)}_{m} (\mathcal{X}^{-}_{k+1})^{(i)} \nonumber \\
       P^-_{k+1} & = \sum_{i=0}^{2L}W^{(i)}_{P} \big[( \mathcal{X}^{-}_{k+1})^{(i)} - \hat{x}^-_{k+1} \big] \big[( \mathcal{X}^{-}_{k+1})^{(i)} - \hat{x}^-_{k+1} \big]^{\sf T} \nonumber\\
       \mathcal{Y}^-_{k+1} & = H( \mathcal{X}^-_{k+1} ) \nonumber \\
       \hat{y}^-_{k+1} & = \sum_{i=0}^{2L} W^{(i)}_{m} (\mathcal{Y}^{-}_{k+1})^{(i)} \nonumber 
   \end{align}
   \STATE Measurement update:
   \begin{align}
       P^{yy}_{k+1} & = \sum_{i=0}^{2L}W^{(i)}_{P} \big[( \mathcal{Y}^{-}_{k+1})^{(i)} - \hat{y}^-_{k+1} \big] \big[( \mathcal{Y}^{-}_{k+1})^{(i)} - \hat{y}^-_{k+1} \big]^{\sf T} \nonumber\\
       P^{xy}_{k+1} & = \sum_{i=0}^{2L}W^{(i)}_{P} \big[(\mathcal{X}^{-}_{k+1})^{(i)} - \hat{x}^-_{k+1} \big] \big[( \mathcal{Y}^{-}_{k+1})^{(i)} - \hat{y}^-_{k+1} \big]^{\sf T} \nonumber\\       
       \hat{x}_{k+1} & = \hat{x}^-_{k+1} + P^{xy}_{k+1} (P^{yy}_{k+1})^{-1} (y_{k+1} - \hat{y}^-_{k+1}) \nonumber \\
       P_{k+1} & = P^-_{k+1} - P^{xy}_{k+1} (P^{yy}_{k+1})^{-1} (P^{xy}_{k+1})^{\sf T} \nonumber
   \end{align}
\end{algorithmic}
\end{algorithm} %

The Unscented Kalman Filter (UKF) process begins with propagating the sigma points in~\eqref{eq:def_sig_pts} using the NN-based dynamics model $F_{\theta}(\cdot)$. The predicted mean $\hat{x}^-_{k+1}$ and covariance $P^-_{k+1}$ in~\eqref{eq:kf_pred} are then approximated through weighted sums of $\mathcal{X}^-_{k+1}$ and state residual covariance, respectively.

Following this, we compute measurements $\mathcal{Y}^-_{k+1}$ by applying the measurement model~\eqref{eq:obs_model} to $\mathcal{X}^-_{k+1}$. The terms $\hat{y}^-_{k+1}$, $P^{yy}_{k+1}$, and $P^{xy}_{k+1}$ in~\eqref{eq:kf_meas_statistics} are approximated using weighted sums of $\mathcal{Y}^-_{k+1}$, measurement residual covariance, and state-measurement residual covariance, respectively.

Finally, we complete the process with the measurement update step described in~\eqref{eq:kf_correct}. \cref{alg:hnn_ukf} presents the complete implementation of the Neural Network-based Unscented Kalman Filter (NN-UKF), which can be integrated with any neural network-based prediction model. We extend this framework by proposing the AHNN-based UKF (AHNN-UKF), which uses AHNN model $F^{(i)}_\theta$ as the prediction model.

\section{Modeling Dynamical Systems} \label{sec:dyn}
We demonstrate the effectiveness of the AHNN and AHNN-UKF through two case studies: mass-spring system (1D) and spacecraft OD on highly elliptic orbits in 2BP with gravitational perturbations (3D).

\subsection{Mass-spring system}
We first consider the dynamic model of the frictionless mass-spring system. The Hamiltonian of the system is described as
\begin{equation}
    \mathcal{H}= \frac{p^2}{2m} + \frac{1}{2} k q^2,
\end{equation}
where $k$ is the spring constant and $m$ is the mass. In the simulation examples, we assume $k=5\text{N/m},\,m=1\text{kg}$.

\subsection{Highly elliptic orbits with gravitational perturbations along a Molniya orbit}
We model spacecraft dynamics in the Two-Body Problem (2BP) setting with gravitational perturbations (per unit mass of spacecraft). The Hamiltonian is given by
\begin{equation}
    \mathcal{H}=\frac{p^2}{2m} + U(q),
\end{equation}
where~\cite{gurfil2016celestial}
\begin{align} \label{eq:grav_potential}
    & U(r,\phi,\theta) =-\frac{\mu m}{r} \bigg[1 
    - \sum^{N_{z}}_{n=2} \bigg(\frac{r_{\rm eq}}{r}\bigg)^{n} J_{n} P^{0}_{n}(\sin \phi) \\
    & + \sum^{N_{t}}_{n=2}\sum^{n}_{m=1} \bigg(\frac{r_{\rm eq}}{r}\bigg)^{n} P^{m}_{n}(\sin \phi)(C^{m}_{n} \cos m\theta + S^{m}_{n} \sin m \theta) \bigg], \nonumber
\end{align}
where $m$, $r=\|q\|$, and $\mu=G M$ denote, respectively, the satellite mass, distance from the center of the primary body, and a gravitational parameter, which is the multiplication of the universal gravitational constant and the primary mass. In~\eqref{eq:grav_potential}, $r_{\rm eq}=6378.1363$ km is the Earth equatorial radius, while $\phi = {\rm arcsin} (z/r)$ and $\theta ={\rm arccos} ( x/r \sqrt{ 1-(z/r)^2 } )$ are the satellite latitude and longitude, respectively, in a planetary body-fixed coordinated frame. The $P^m_n$ are the associated Legendre functions of degree $n$ and order $m$, and $C^m_n,\; S^m_n$ are spherical harmonics coefficients of degree $n$ and order $m$. The value of the gravitational parameter for Earth orbits is $\mu=398600.4418$ km$^2$/s$^2$, which corresponds to the gravitational parameter of the Earth. It is sufficient to set $m=1$ kg (as satellites with different masses will follow the same orbit for the model considered here). Note that ($q,p$) are expressed in the Earth-centered Inertial frame (or in a similarly defined planet-centric inertial frame associated with other planets). 
\begin{figure}[ht!]
\vskip -0.1in
\begin{center}
\centerline{\includegraphics[width=.65\columnwidth]{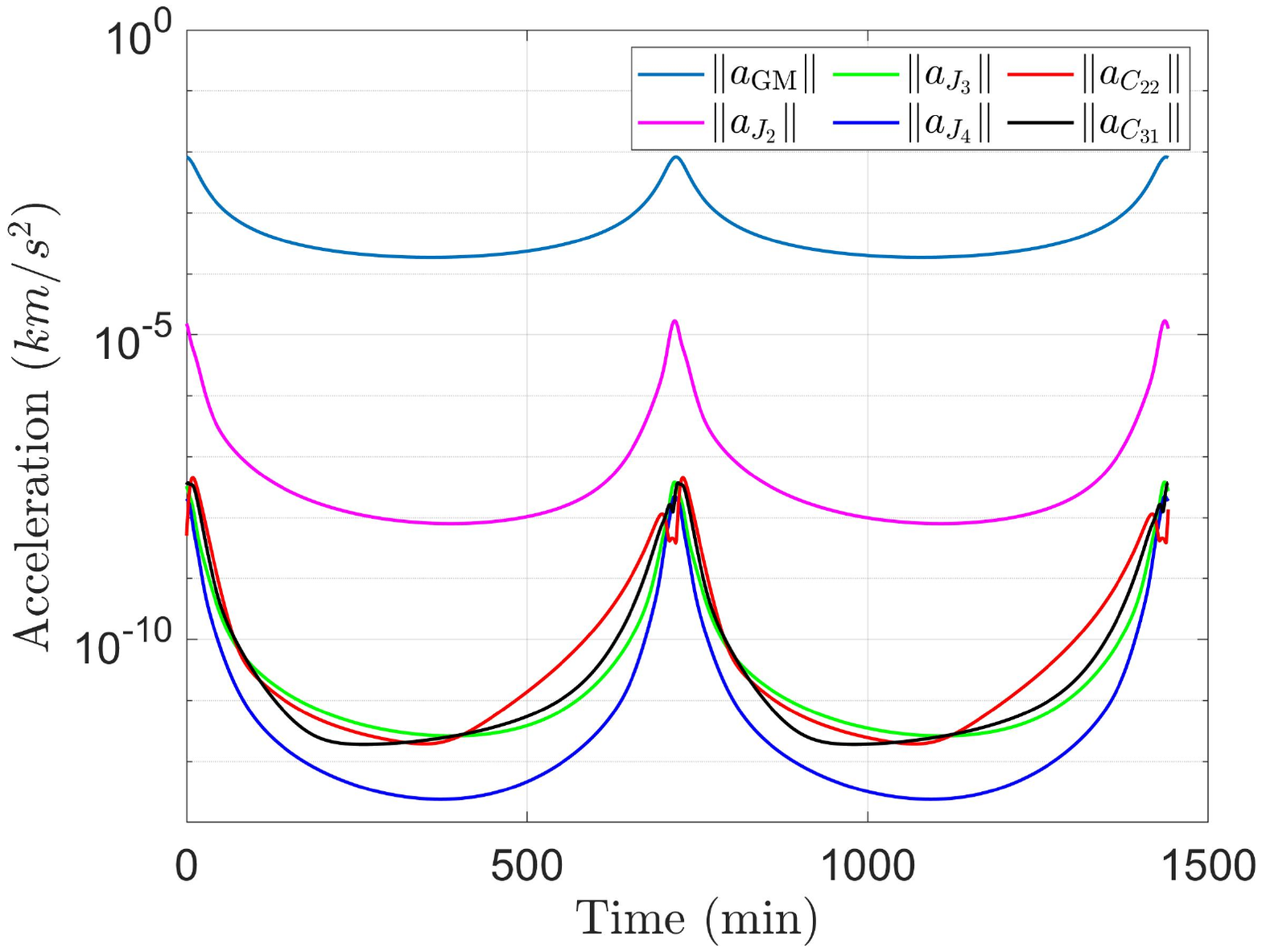}}
\caption{Accelerations from gravitational perturbations along the sample highly elliptic orbit.}
\label{fig:molniya_pert}
\end{center}
\vskip -0.1in
\end{figure}

Figure~\ref{fig:molniya_pert} shows the acceleration magnitude evolution over the reference highly elliptic orbit. In this paper, we consider the $J_2$ effect with the coefficient $J_2=1.0826 \times 10^{-3}$ in~\eqref{eq:grav_potential} since the $J_2$ effect is the dominant perturbed acceleration while others are negligible compared to that from the point mass gravity acceleration $\|a_{\rm GM}\|$. Note that the GGM05C mean gravity field model~\cite{cheng2011variations} provides the necessary gravitational coefficients.

\section{Experiments} \label{sec:exp}
We present numerical experiments to demonstrate the performance of our proposed method in two prediction case studies for a spring-mass system (1D motion)
and highly elliptic orbits (3D motion), informed by the higher-order gravitational potential, respectively. We evaluate prediction performance under two scenarios: one assuming known true states and the other assuming perturbed states with noisy measurements. Additionally, we present ablation studies of window size, $W\in\{1,3,5\}$, in~\eqref{eq:loss_ahnn}. 

\subsection{Data preparation}
In order to train and evaluate models, we employed a hold-out method, splitting the data into training ($80\%$), validation ($15\%$), and test ($5\%$) sets. The training data is normalized by min-max normalization. 

\subsubsection{Mass-spring system}

We generated $2$,$500$ trajectories using initial conditions sampled uniformly from $[-1, 1]^2$ for both position and momentum coordinates. Each trajectory was numerically integrated using a 4th-order Gauss-Legendre (GL4) symplectic integrator \cite{bogaert2014iteration} with fixed-point iteration. The integration was performed over the time interval $[0, 10]$ s and trajectories were sampled every $0.01$ s. 

\subsubsection{ Elliptic Orbits}

For the elliptic orbit, we sampled $2$,$500$ initial position and velocity pairs from the uniform distribution for initial altitude at periapsis between [$540,560$] km, orbit eccentricity between [$0.7, 0.8$] and inclination between [$60^\circ, 66^\circ$]. The trajectories were simulated for two periods of over 1 day and sampled every $60$ s. In the training data set, each sample orbit can have an altitude between [$540,49$,$000$] km. The large differences in altitudes introduce additional challenges when we train the models. 

For generating the training data, we use `KahanLi8'~\cite{kahan1997composition}, which is the 8th order explicit symplectic integrator, in `\texttt{DifferentialEquations.jl}' package. The `KahanLi8' was chosen as it is known to compute accurate solutions for Hamiltonian systems~\cite{clain2025structuralschemeshamiltoniansystems}.

\subsection{Training \& Evaluation}
We conduct experiments using NVIDIA 3080 Ti GPU. We use a fourth-order Runge-Kutta integrator with a fixed time step of $60$ s in~\eqref{eq:hnn_pred} and Figure~\ref{fig:hnn}.

We update learnable parameters using AdamW Optimizer~\cite{loshchilov2017decoupled} ($\beta_1=0.9$, $\beta_2=0.999$, $\epsilon=10^{-8}$, $\lambda=0.01$) in PyTorch~\cite{paszke2019pytorch} with the batch size of $256$ and `\textit{ExpHyperbolicLR}' learning rate scheduler~\cite{kim2024hyperboliclr}.
For HNN and MLP, we search over network configurations with $3$-$10$ layers and ${128,256,512,1024}$ nodes, learning rates in the interval $[10^{-4},10^{-3}]$, scheduler parameters with upper bound in the interval $[300,400]$ and infimum learning rate in the interval $[10^{-7},10^{-4}]$. For NODE, due to its sensitivity, we modify the search range of learning rates to the interval $[10^{-6},10^{-4}]$ and infimum learning rate to the interval $[10^{-8},10^{-6}]$ while maintaining other configurations. We train our model for $250$ epochs using the optimal hyperparameter set. 

For efficient training, we introduce the Predicted Final Loss (PFL) pruner, which forecasts the validation loss at epoch 50 using an exponential model fitting based on the first 10 epochs' loss curve. Early stopping is applied if the predicted loss exceeds the final validation loss of the top 10 performing models. In order to generalize model performance, we evaluate the each model with three random seeds, [$7,201,719$]. 

For UKF implementation, we optimize hyperparameters ($\alpha,\beta,\kappa$) over $200$ trials to minimize the root mean squared error (RMSE) between the true and predicted position and velocity using the TPE algorithm, while setting the measurement update frequency as $60$ steps.  
(Average 10 orbit's RMSE - orbits are from validation set)

\subsection{Results}
We report experimental results in terms of the RMSE of the predicted state and Hamiltonian trajectories with respect to that of true values, respectively, and an illustration of the AHNN$_5$KF prediction. We also report the comparison of our proposed methods with other baseline NN-based models, including MLP, NODE, and HNN, highlighting the effectiveness of our proposed methods.

\subsubsection{Mass-spring system}
\begin{figure}[ht]
\vskip 0.1in
\begin{center}
\subfigure{
    \includegraphics[width=0.47\columnwidth]{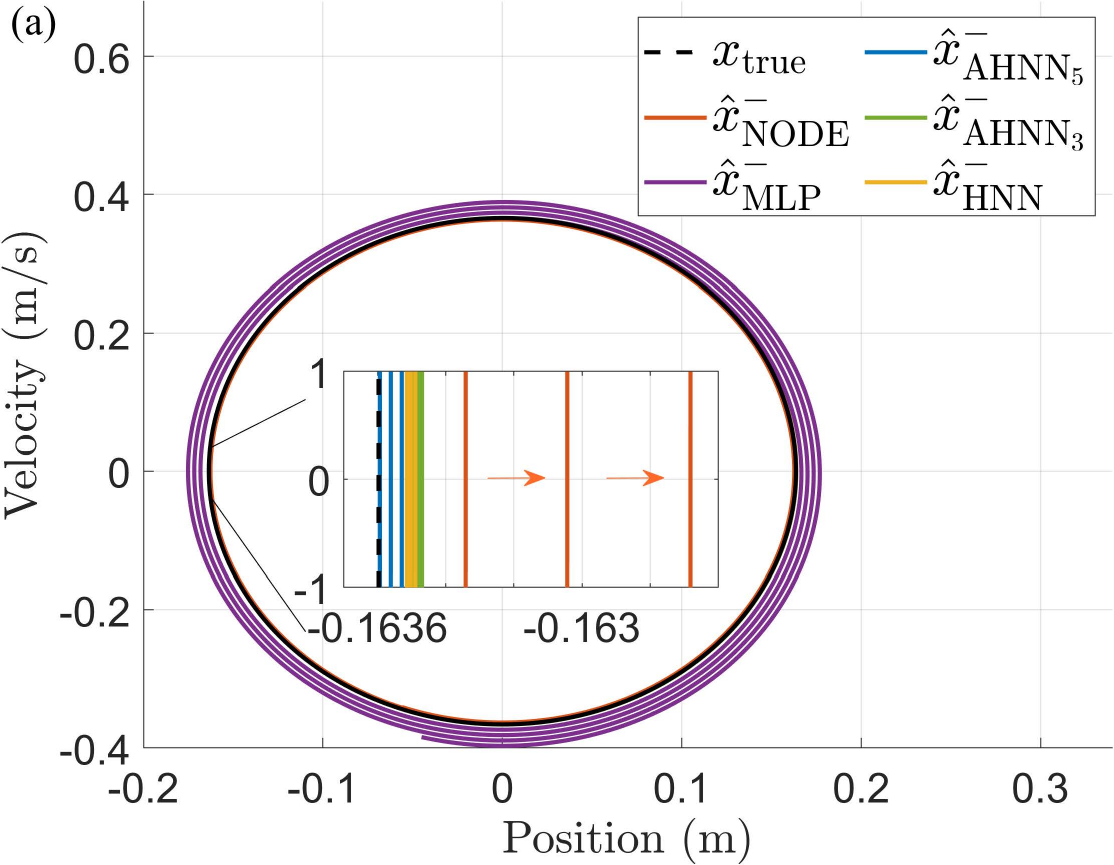}
}
\subfigure{
    \includegraphics[width=0.47\columnwidth]{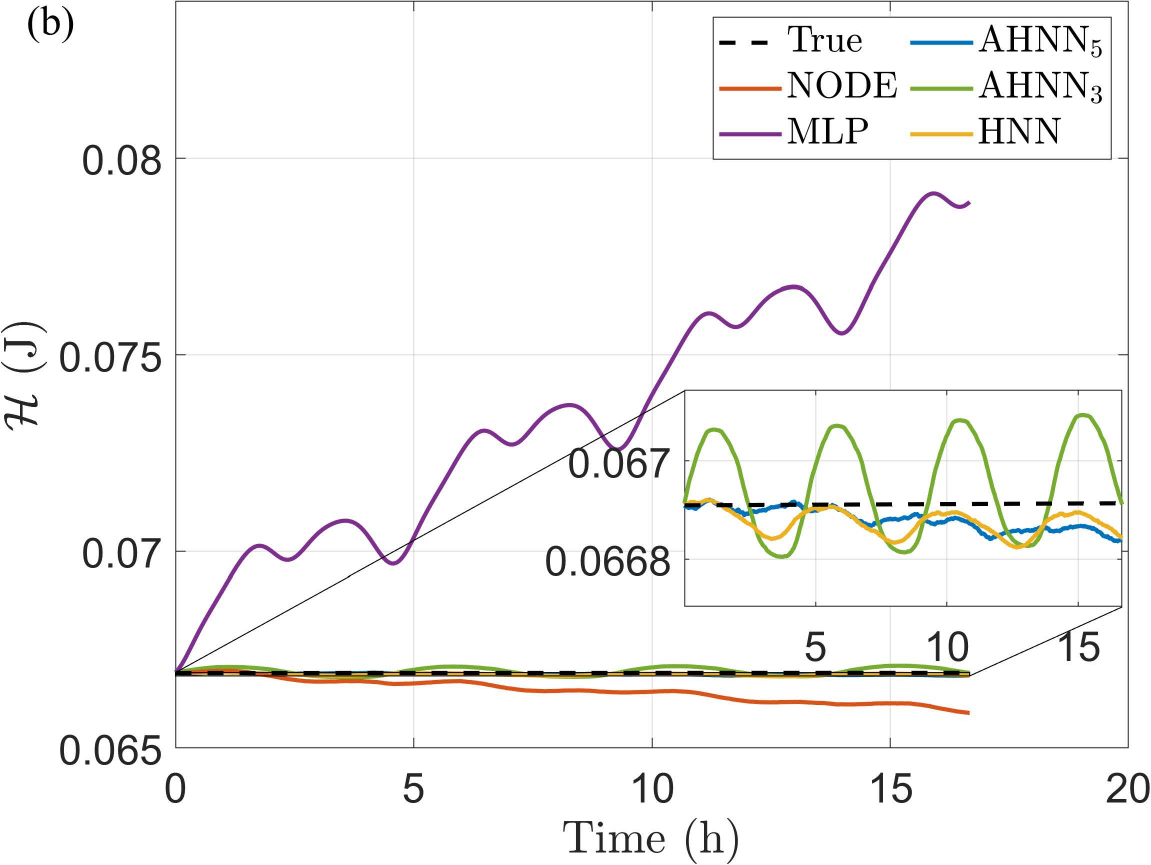}
}
\caption{(a) The state prediction of the simple mass-spring system. (b) The energy evolution over time. }
\label{fig:pred_energy_mass}
\end{center}
\vskip -0.1in
\end{figure}
Figure~\ref{fig:pred_energy_mass} compares predicted trajectories and energy evolution over time for implementations without UKF. In the phase space, MLP predictions show trajectories with an expanding radius, while NODE predictions exhibit a contracting radius compared to ground truth. Consequently, the system energy increases with MLP and decreases with NODE, as shown in Figure~\ref{fig:pred_energy_mass} (b). Using the units of J, RMSE values of calculated energies are 7.211$\times 10^{-3}$ with MLP, 5.2546$\times 10^{-4}$ with NODE, 4.4720$\times 10^{-5}$ with HNN, 1.0630$\times 10^{-4}$ with AHNN$_3$, and 3.8400$\times 10^{-5}$ with AHNN$_5$. Thus, HNN, AHNN$_3$, and AHNN$_5$ show nearly constant energy levels close to the initial value, while achieving the lowest prediction error with AHNN$_5$ as stated in Table~\ref{tab:rmse_sho}.

Table~\ref{tab:rmse_sho} compares prediction accuracies for the frictionless mass-spring system using various NN-based methods with and without UKF.
The state prediction with NODE is 15 times more accurate in terms of RMSE than MLP when the prediction starts at the true value.
The HNN, then, computes more accurate predictions than NODE, successfully replicating the ideal pendulum case study in~\cite{greydanus2019hamiltonian}.
Since the complexity of the mass-spring system is low, the prediction accuracy of AHNN$_3$ is comparable with that of HNN, but the prediction accuracy improvements become more apparent when AHNN$_5$ is considered.
\begin{table}[ht]
\caption{RMSE for the mass-spring system.}
\vskip 0.1in
\begin{center}
\begin{small}
\begin{sc}
\begin{tabular}{c|cccc}
    \toprule
    \multirow{2}{*}{\makecell{Model \\ (10$^{-4}$, 10$^{-4}$)}} & \multicolumn{2}{c}{True Initial}  & \multicolumn{2}{c}{Perturbed Initial} \\
      & Pos & Vel & Pos & Vel \\
    \midrule
    MLP      & $91.730$       & $205.09$       & $92.187$       & $205.89$       \\
    NODE     & $6.3214$       & $13.659$       & $9.5715$       & $29.967$       \\    
    HNN      & $3.1322$       & $7.2384$       & $7.1945$       & $16.132$       \\
    AHNN$_3$ & $4.0656$       & $8.3772$       & $8.6479$       & $18.877$       \\    
    AHNN$_5$ & $\pmb{1.5528}$ & $\pmb{3.5681}$ & $\pmb{6.5678}$ & $\pmb{14.586}$ \\    
    \midrule
    MLPKF      & $15.559$       & $49.045$        & $15.642$       & $49.205$        \\    
    NODEKF     & $5.6026$       & $1.2286$        & $1.7998$       & $4.5569$        \\    
    HNNKF      & $0.7261$       & $2.9779$        & $1.5580$       & $4.2539$        \\    
    AHNNKF$_3$ & $1.2275$       & $3.7587$        & $1.8741$       & $4.9310$        \\    
    AHNNKF$_5$ & $\pmb{0.6596}$ &  $\pmb{2.2451}$ & $\pmb{1.5166}$ & $\pmb{3.7685}$  \\    
    \bottomrule
\end{tabular}
\end{sc}
\end{small}
\end{center}
\vskip -0.1in
\label{tab:rmse_sho}
\end{table}
\begin{figure}[ht]
\vskip 0.1in
\begin{center}
\centerline{\includegraphics[width=0.95\columnwidth]{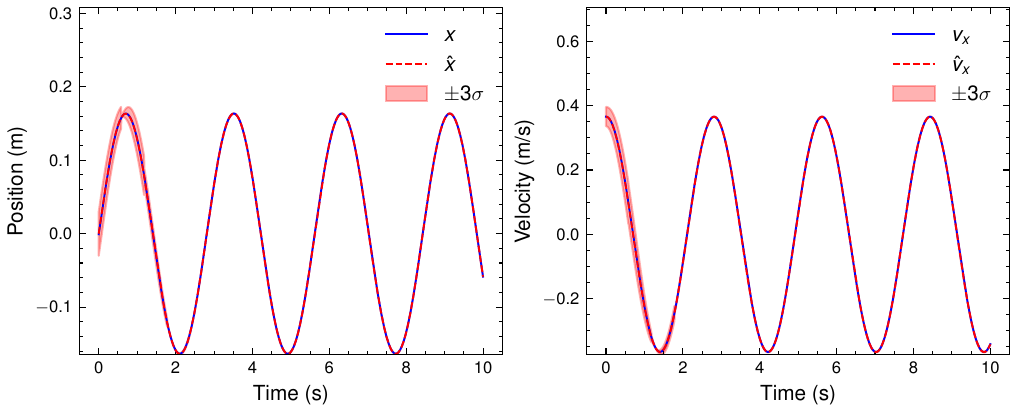}}
\caption{Mass-spring system prediction results of AHNN$_5$ with UKF integration. Predictions are made with 60-step observation frequency and perturbed initial conditions.}
\label{fig:mass_spring_60}
\end{center}
\vskip -0.1in
\end{figure}
Figure~\ref{fig:mass_spring_60} demonstrates AHNNKF$_5$ performance with a perturbed initial state and covariance estimate of $P_0= 10^{-7} \mathbf{I}_2$ and position measurements. The UKF integration reduces state uncertainty (shown as red shading) and corrects the perturbed initial state prediction toward the true trajectory using measurement data.

Table~\ref{tab:rmse_sho} shows that UKF integration improves RMSE for all neural network models, with AHNNKF$_5$ achieving the best performance. The UKF effectively handles both model mismatch errors when starting from true initial states and combined errors from initial perturbations and model mismatch when starting from perturbed states, while providing uncertainty quantification.

\subsubsection{Elliptic orbits}
\begin{figure}[ht!]
\vskip 0.1in
\begin{center}
\centerline{\includegraphics[width=1.8 in]{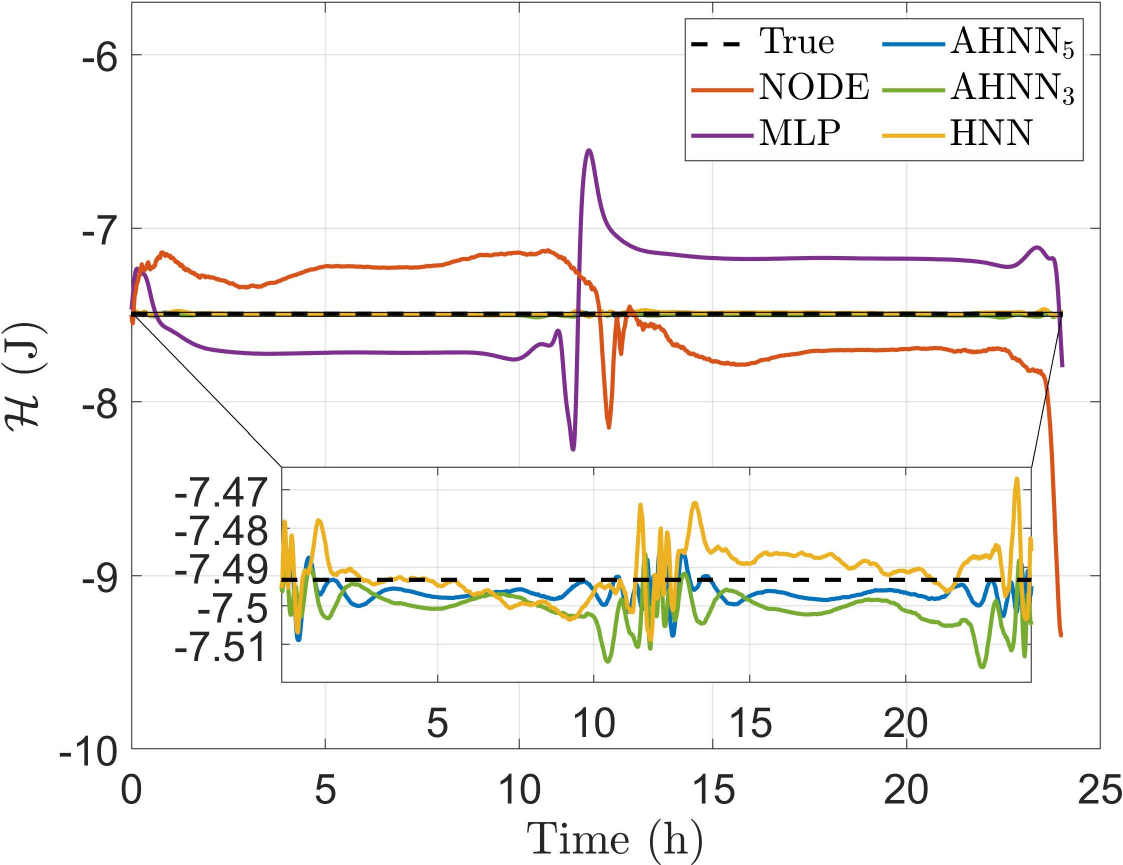}}
\caption{The energy evolution of object along a highly elliptic orbit over time.}
\label{fig:molniya_energy}
\end{center}
\vskip -0.1in
\end{figure}
Figure~\ref{fig:elliptic_orbit} presents a graphical comparison of predicted trajectories between our approach without UKF and other NN models. Unlike predicted trajectories of HNN, AHNN$_3$, and AHNN$_5$, MLP predicts a trajectory with an increasing orbit radius, and NODE predicts a trajectory with a gradually decaying orbit radius. Table~\ref{tab:rmse_elliptic} reports prediction accuracies for highly elliptic orbits using various NN models. The RMSE values of predicted states using HNN decrease by an order of magnitude compared to MLP and NODE. Additionally, the RMSE values of AHNN$_5$ predictions indicate a one-order reduction compared to that of HNN.

\begin{table}[ht]
\caption{RMSE for the highly elliptic orbit task.}
\vskip 0.1in
\begin{center}
\begin{small}
\begin{sc}
\begin{tabular}{c|cccc}
    \toprule
    \multirow{2}{*}{\makecell{Model \\ (10$^{2}$,10$^{-1}$)}} & \multicolumn{2}{c}{True Initial}  & \multicolumn{2}{c}{Perturbed Initial} \\
      & Pos & Vel & Pos & Vel \\
    \midrule    
    MLP      & $153.03$       & $33.635$       & $190.81$       & $422.97$ \\
    NODE     & $124.66$       & $84.067$       & $ 14374$       & $ 25121$ \\    
    HNN      & $17.330$       & $13.516$       & $92.421$       & $334.44$ \\
    AHNN$_3$ & $2.8592$       & $0.6476$       & $90.122$       & $219.06$  \\    
    AHNN$_5$ & $\pmb{1.7381}$ & $\pmb{4.5962}$ & $\pmb{89.676}$ & $\pmb{218.67}$  \\
    \midrule
    MLPKF      & $4.8917$       & $2.0256$       & $2.1175$       & $1.1613$       \\    
    NODEKF     & $1.0257$       & $0.6056$       & $0.9615$       & $0.5616$       \\  
    HNNKF      & $4.7009$       & $41.683$       & $11.672$       & $60.738$       \\    
    AHNNKF$_3$ & $0.4235$       & $0.2440$       & $0.5286$       & $0.2865$       \\    
    AHNNKF$_5$ & $\pmb{0.3689}$ & $\pmb{0.2119}$ & $\pmb{0.5038}$ & $\pmb{0.2501}$ \\    
    \bottomrule
\end{tabular}
\end{sc}
\end{small}
\end{center}
\vskip -0.1in
\label{tab:rmse_elliptic}
\end{table}

In Figure~\ref{fig:molniya_energy}, the energies of predicted trajectories with MLP and NODE indicate a significant lack of energy conservation at the closest point to the primary body's center, where the velocity of a satellite rapidly changes. Contrarily, HNN, AHNN$_3$, and AHNN$_5$ conserve the initial energy and show robustness to large velocity changes, although small perturbations appear at the closest point to the primary body's center. 
Using the units of J, RMSE values of calculated energies are 0.3020 with MLP, 0.3091 with NODE, 6.4312$\times 10^{-3}$ with HNN, 8.5515$\times 10^{-3}$ with AHNN$_3$, and 4.3444$\times 10^{-3}$ with AHNN$_5$, indicating that the capabilities of energy conservation of HNN, AHNN$_3$, and AHNN$_5$, outperform by showing a two-order reduction compared to MLP and NODE, but their performances in energy conservation are comparable each other.

\begin{figure}[ht]
\vskip 0.1in
\begin{center}
\centerline{\includegraphics[width=0.95\columnwidth]{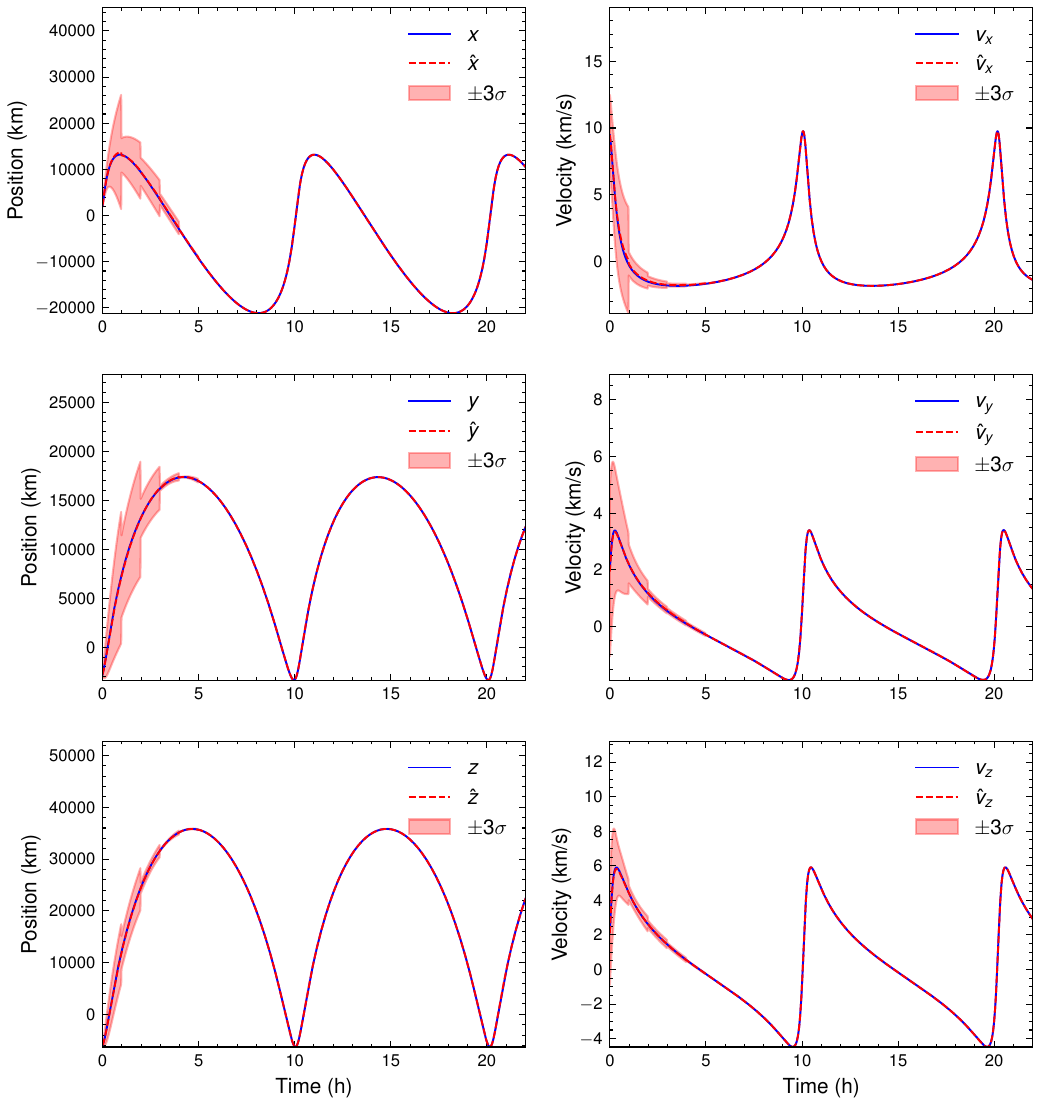}}
\caption{Orbital prediction results of AHNN$_5$ with UKF integration for elliptic orbits. Predictions are made with 60-step observation frequency and perturbed initial conditions.}
\label{fig:molniya_AHNNKF_60}
\end{center}
\vskip -0.1in
\end{figure}

Figure~\ref{fig:molniya_AHNNKF_60} demonstrates AHNNKF$_5$ performance with a perturbed initial state and covariance estimate of $P_0={\rm diag}(10,10,10^{-3},10^{-3})$. The UKF integration reduces state uncertainty (shown as red shading) and corrects the perturbed initial state prediction toward the true trajectory using measurement data.

Table~\ref{tab:rmse_elliptic} shows that Kalman Filter integration improves RMSE for all neural network models, with AHNNKF$_5$ achieving the best performance. The UKF effectively handles both model mismatch errors when starting from true initial states and combined errors from initial perturbations and model mismatch when starting from perturbed states, while providing uncertainty quantification.

It is worthwhile to note that NODE indicates a lack of robustness when faced with the perturbed initial states, as shown in Table~\ref{tab:rmse_elliptic}. While combining NODE with UKF improves prediction accuracy, the NODEKF's RMSE values often blow up during hyperparameter optimization ($\alpha,\beta,\kappa$), highlighting the robustness of UKF with HNN and AHNNs.

\section{Conclusion} \label{sec:conclusion}
This paper presents an effective approach for predicting Hamiltonian dynamical systems using neural networks and Bayesian data assimilation. Our proposed Autoregressive Hamiltonian Neural Network (AHNN) demonstrates improved long-term prediction accuracy while preserving physical invariants, e.g., energy. Integrating the unscented Kalman filter with AHNN enables real-time refinement of predictions using online measurement data and uncertainty quantification. Through numerical experiments on the simple mass-spring system and on the highly elliptic orbits in the gravitational system, we validate that our method successfully captures highly nonlinear dynamics and provides accurate and robust predictions of true state and of perturbed state. The results suggest promising applications in spacecraft trajectory prediction and celestial mechanics, where accurate long-term forecasting is essential. This method can be applied to other Hamiltonian systems; however, combining our approach with various dynamical models and comparing their performance left future work. 

{\color{red} }

\section*{Acknowledgements}
This research is supported by the Air Force Office of Scientific Research Grant number FA9550-23-1-0678.

\bibliography{reference}
\bibliographystyle{icml2025}

\newpage
\appendix
\onecolumn
\section{Analysis of Error Dynamics} \label{sec:app1}
To visualize RMSE evolution over time, we employed a Simple Moving Average (SMA). The SMA alleviates high-frequency variations in the RMSE values while maintaining the underlying patterns. The SMA computes the unweighted mean of the past $n$ time series data points~\cite{brown2004smoothing}. Let $X(t_k)=\left(x(t_i)\right)_{i=j}^{k}$, for $0\leq j \leq k$ represent a sequence of data points from the $i$-th to $j$-th data point. The SMA $\bar{X}(t_k)$ for the data sequence $X(t_k)$ is defined as
\begin{equation} \label{eq:simple_moving_avg}
    \bar{X}(t_k) = \begin{cases}
        \frac{1}{n}\sum_{i=k-n+1}^{k} x_i, & \text{for } k \geq n, \\
        \frac{1}{k}\sum_{i=1}^{k} x_i, & \text{for } k < n,
    \end{cases}    
\end{equation}
where $n=k-j$ denotes a predetermined window size. We selected the window size of $240$ steps, i.e., $n=240$. The sampling time was set to $0.01$ s for the mass-spring system and $60$ s for elliptic orbit simulations. While selecting an appropriate window size is important for achieving proper smoothing to reveal underlying trends, this consideration lies beyond the scope of this paper.

\subsection{Mass-spring system}
\begin{figure}[ht]
\vskip 0.1in
\begin{center}
\centerline{\includegraphics[width=.55\columnwidth]{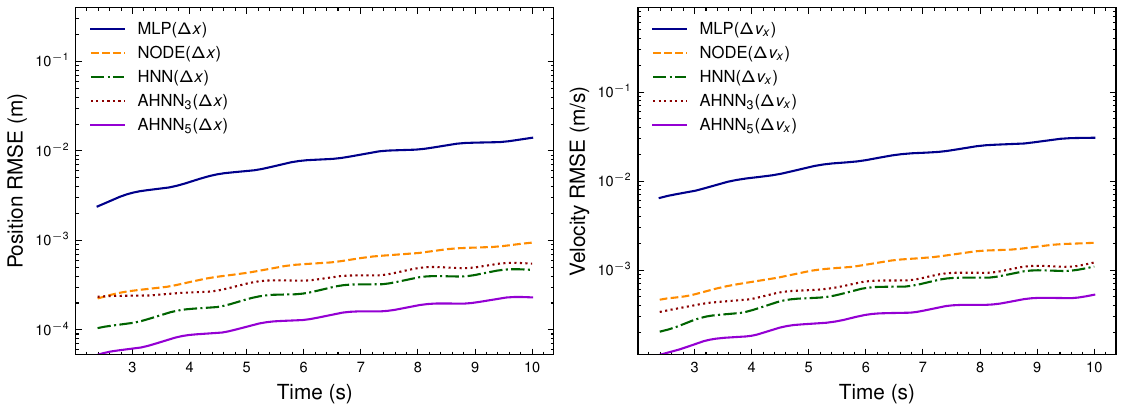}}
\caption{The SMAs of position and velocity RMSEs with a 240-step window using different NN models based on 125 test trajectories for the mass-spring system.}
\label{fig:moving_rmse_sho}
\end{center}
\vskip -0.1in
\end{figure}

Figure~\ref{fig:moving_rmse_sho} compares the SMA values for the frictionless mass-spring system using various NN-based methods without UKF. For all methods, the accumulated prediction errors lead to increases in SMA values over time. The SMA trajectory with NODE is ten times smaller than that with MLP, but twice larger than that with HNN. The SMA value with AHNN$_3$ initially is larger than that with HNN, and they become comparable at the end. AHNN$_5$ presents the smallest SMA over the experiment, highlighting the best prediction performance in our ablation study.

\begin{figure}[ht]
\vskip 0.1in
\begin{center}
\centerline{\includegraphics[width=.55\columnwidth]{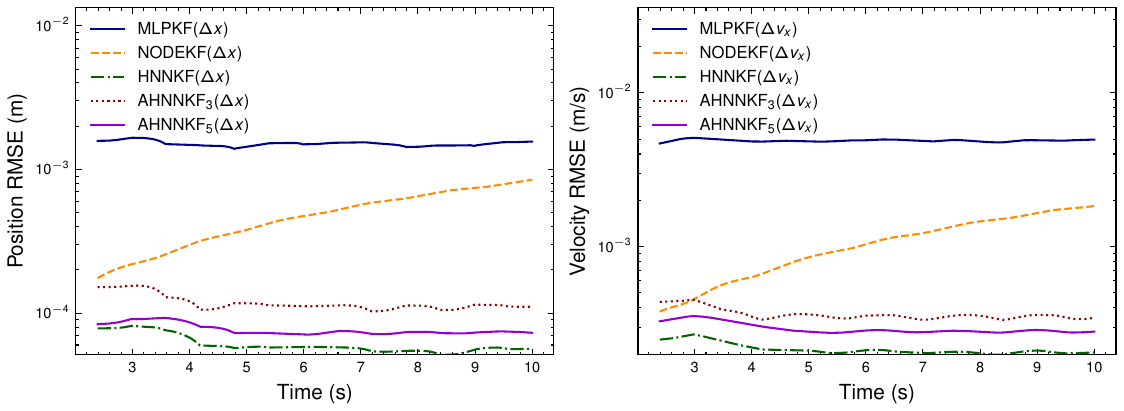}}
\caption{The SMAs of position and velocity RMSEs with a 240-step window using different NN models with UKF based on 125 test trajectories for the mass-spring system.}
\label{fig:moving_rmse_sho_60}
\end{center}
\vskip -0.1in
\end{figure}
Figure~\ref{fig:moving_rmse_sho_60} reports the SMA values using different NN-based UKF in the mass-spring system. Integrating NN models with UKF improves prediction accuracies in terms of SMA values. Except for NODEKF, other NN-based KFs improve prediction accuracies and handle the NN model prediction error by assimilating online measurement data. The NODEKF, however, does not show clear improvements in SMA values. This is because NODE is sensitive with respect to perturbation. Therefore, NN models with UKF demonstrate their effectiveness in long-term prediction, but NODE requires further adjustment.

The prediction accuracies between HNN and AHNN variants, e.g., AHNN$_3$ and AHNN$_5$, are comparable with and without UKF. The experimental result for the mass-spring system concludes that the standard HNN is able to learn Hamiltonian and to compute accurate predictions, while additional loss function for AHNN complicates the training process.

\subsection{Elliptic orbit}

\begin{figure}[!t]
\vskip 0.1in
\begin{center}
\centerline{\includegraphics[width=0.8\columnwidth]{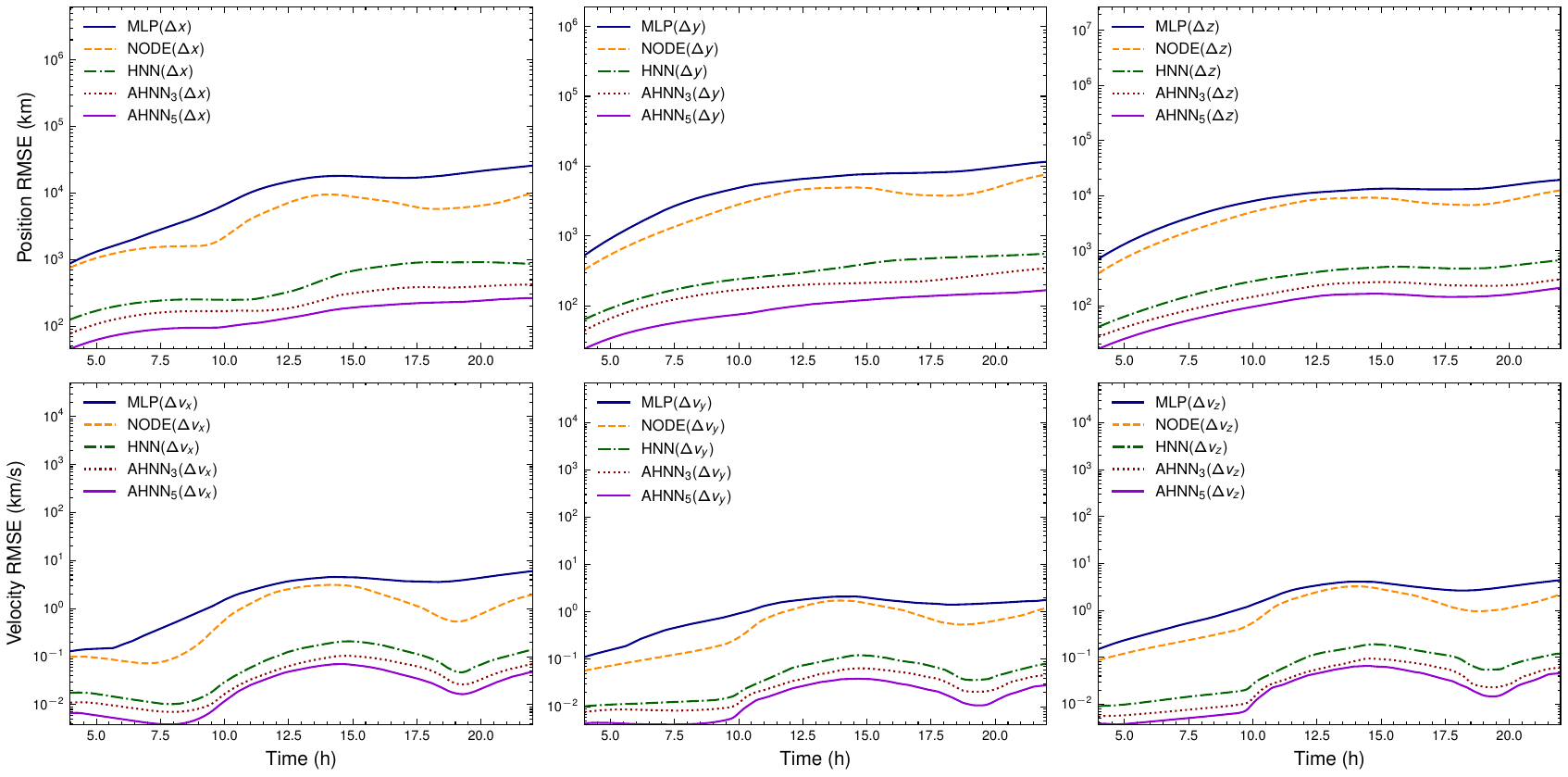}}
\caption{SMA of position and velocity RMSE with a 240-step window for different neural network architectures in elliptic orbit. Position RMSE (km) and velocity RMSE (km/s) are shown for $x$, $y$, and $z$ components, computed from 125 test orbits.}
\label{fig:moving_rmse_molniya}
\end{center}
\vskip -0.1in
\end{figure}
\begin{figure}[!t]
\vskip 0.1in
\begin{center}
\centerline{\includegraphics[width=0.8\columnwidth]{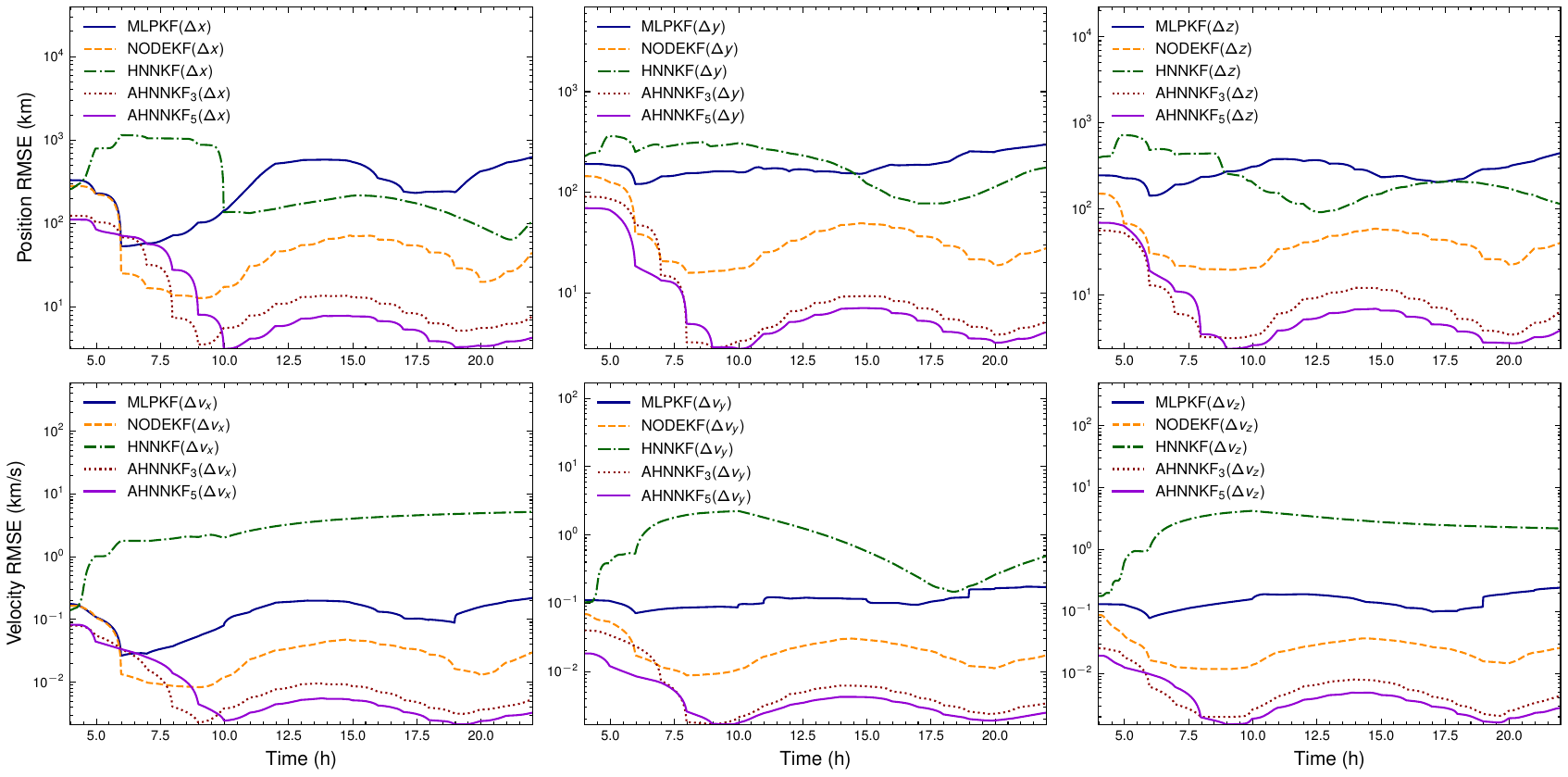}}
\caption{SMA of position and velocity RMSE with a 240-step window for different neural network architectures with UKF in elliptic orbit. Position RMSE (km) and velocity RMSE (km/s) are shown for $x$, $y$, and $z$ components, computed from 125 test orbits with 60-step observation frequency.}
\label{fig:moving_rmse_molniya_60}
\end{center}
\vskip -0.1in
\end{figure}

Figure~\ref{fig:moving_rmse_molniya} presents SMA values for elliptic orbit prediction using various NN models without UKF.
The cumulative prediction errors are shown as increases in SMA values over time. In this complicated system, predictions using HNN and AHNN variants improve an order of magnitude in SMA values, bolstering the importance of energy conservation. Compared to HNN, AHNN variants then present additional improvement in prediction accuracy.

Figure~\ref{fig:moving_rmse_molniya_60} describes the SMA values with various NN-based UKF for spacecraft orbit determination on highly elliptic orbits. The MLP, NODE, and AHNN variants advance prediction accuracy by integrating UKF, while HNN does not. In particular, AHNNKFs outperform all other NN-based KF, showing about two orders of magnitude improvement in prediction accuracy compared to HNNKF. Therefore, learning accurate dynamic models poses a huge challenge in the spacecraft orbit determination, despite the support of online measurement data.

\end{document}